# Multi-Task Generative Adversarial Nets with Shared Memory for Cross-Domain Coordination Control

JunPing Wang, WenSheng Zhang, Ian Thomas, ShiHui Duan, YouKang Shi

*Abstract*—Generating sequential decision process from huge amounts of measured process data is a future research direction for collaborative factory automation, making full use of those online or offline process data to directly design flexible make decisions policy, and evaluate performance. The key challenges for the sequential decision process is to online generate sequential decision-making policy directly, and transferring knowledge across tasks domain. Most multi-task policy generating algorithms often suffer from insufficient generating cross-task sharing structure at discrete-time nonlinear systems with applications. This paper proposes the multi-task generative adversarial nets with shared memory for cross-domain coordination control, which can generate sequential decision policy directly from raw sensory input of all of tasks, and online evaluate performance of system actions in discrete-time nonlinear systems. Experiments have been undertaken using a professional flexible manufacturing testbed deployed within a smart factory of Weichai Power in China. Results on three groups of discrete-time nonlinear control tasks show that our proposed model can availably improve the performance of task with the help of other related tasks.

## I. INTRODUCTION

THE distributed coordination control (DCC) has gained great interest in the past years, partly due to its wide applications in many industrial systems, including flexibility manufacturing, multiple robot systems, unmanned aerial vehicles, as well as power networks [1]. The DCC will enable large scale autonomous systems to dynamically coordinate their actions across task domain. All autonomous systems operating as part of the DCC will run in an iterative three-step closed loop: learning policies directly from the environment, performing internal computations on the data and responding by performing actions that affect the environment - either by altering operational status or via communication with other objects [2]. Two levels of the DCC are associated with this loop: deliberative and meta-level. Meta-level control provides overall coordination of the resources in each autonomous system. Two levels of the DCC offer many advantages in terms of flexibility, reliability, manipulability, and scalability that cannot be achieved by an individual system [3]. Huge amounts of measured process data can easily be collected by the DCC, both in the form of stored historical data from prior measurements and online data in real time during process runs. Thus, it is very significant if we can make full use of massive online or offline process data to directly generate sequential decision policies model, and online transfer those policies model across multi-task domain [4].

Most previous multi-task reinforcement learning algorithm have shown considerable recent success in solving high-dimensional sequential decision-making problem, transferring knowledge between tasks to accelerate learning [5]. But the training cost of this algorithm is very prohibitively expensive, especially in scenarios where an autonomous system will face multiple tasks and must be able to quickly learn control policies for each new task. Recently, significant progress has been made through combining advances in training deep neural networks with reinforcement learning, termed a Deep Q-Network (DQN) algorithm that can learn successful sequential decision policies directly from high-dimensional sensory inputs, and approximate the optimal action-value function. Those algorithm in deep learning have only involved discriminative models, usually which can learn successful policies directly from high-dimensional sensory inputs using end-to-end reinforcement learning [6]. However, those algorithms have the difficulty of approximating many intractable probabilistic computations in maximum likelihood estimation,and leveraging the benefits of transferring generative sequence accurately between tasks.

Generative adversarial nets (GAN) [7] is successful deep generative framework for solving the above problem via two player minimax game with value function $V(D,G)$. Specifically, one discriminative model $D$ in the GAN alternately guides the training of the generative model from real-valued data, and the other generative model $G$ simultaneously learns to enhance recognizing capability of discriminative model by generating high quality data. This approach has been successful and been mostly applied in computer vision tasks of generating samples of natural images [8]. Unfortunately, applying GAN to generating sequential decision policies model has two problems in multi-task learning setting. Firstly, when the GAN generate sequential deterministic policies directly from raw sensory input of all of tasks, how to approximate the optimal action-value function $Q^*(s,a)$ become crucial problem over continuous action spaces by the discriminative model. Secondly, it is importance problem that incorporates a shared

J. Wang and W. Zhang are with Laboratory of Precision Sensing and Control Center, Institute of Automation, Chinese Academy, Beijing, China Email: wangjunping@bupt.edu.cn, wensheng.zhang@ia.ac.cn.

I. Thomas is with CHIEF Technology Officer, FUJITSU, France, Email: ian@runmyprocess.com.

S. Duan and Y. Shi are with Communications Standards Research Institute, China Academy of Telecommunication Research of MIIT, Beijing, China Email: duanshihui@ritt.cn, shiyoukang@ritt.cn.

discrete tokens memory into multi-task policy gradient to enable transfer between several related tasks. There are currently no general methods to learn an inter-task mapping without requiring either current tokens in multi-task policy gradient settings, or an expensive analysis of an exponential number of inter-task mappings in the size of the state and action spaces.

In order to address above two challenges, this paper proposes multi-task generative adversarial nets with shared memory for big data-driven cross-domain coordination control, in which the framework simultaneously train two model: one actor network generator that is capable to generate successful sequential decision policies model directly from raw sensory input of all of tasks by DQN algorithm , another critic network discriminator that estimates the large-scale action-value function $Q^*(s,a)$ from real sequence data or the generator. The training procedure for the actor network generator is to maximize cumulative future reward of critic network discriminator. When each autonomous system faces multiple tasks, an autonomous system in the coordination control framework will run in a three-step closed loop: actor network generator, critic network discriminator, and selecting optimal joint action. The main contributions are as follows.

1) The cross-domain coordination control(CDCC) for collaborative factory automation which making full use of those online or offline process data to design flexible make decisions policy directly, evaluate performance of system actions, perform real-time optimization, and conduct fault diagnosis. The aim of the CDCC is to address cross-domain sequential decision problem of large scale autonomous system on two different levels: critic network layer, and actor network layer.

2) The multi-task generative adversarial nets with shared memory (MTGAN) is developed to generate sequential decision policy directly from raw sensory input of all of tasks, online evaluate performance of system actions, and transfer cross-task sharing structure at every layer in discrete-time nonlinear systems to accelerate learning. The MTGAN facilitate iterative three-step closed loop in collaborative factory automation by actor-critic adversarial process, in which simultaneously train two models: actor network generator, critic network discriminator.

3) This paper apply CDCC and MTGAN algorithm into the three dynamical system with professional flexible manufacturing testbed of Weichai Power in China. In each control system, the task distance between the current state and the goal position was used as the reward function. On all systems, the reward function is based on two factors: a) penalizing states far from the goal state, and b) penalizing actions to encourage smooth, low-energy manufacturing.

The remainder of the paper is organized as follows: Section II presents related works covering data-driven control and learning systems in cross-domain dynamical cooperation. Section III describes the details of the multi-task generative adversarial nets algorithm with shared memory for crossdomain dynamical cooperation, while Section IV describes the experiment scenarios and results. Section V finally concludes the paper.

## II. RELATED WORKS AND MOTIVATION

Distributed coordination control has attracted extensive research interests in flexibility manufacturing, multiple robot systems, sensor networks, unmanned aerial vehicles, and power networks. Through analyzing very important progresses about distributed coordination control in the past five years, multi-task deep reinforcement learning is a significant future researching topic for distributed coordination control.

The multiagent reinforcement learning in [9] provides a common approach that allows agents to interact with stochastic environment and receive rewards in the distributed coordination control. Several multiagent reinforcement learning algorithms have been proposed in [10], all of which have some theoretical results of convergence in general-sum games. A common assumption of these algorithms is that an player knows its own payoff matrix. To guarantee convergence, each algorithm has its own additional assumptions, such as requiring a player to know a Nash Equilibrium and the strategy of other players, or observe what actions other players executed and what rewards they received [11]. However, For practical applications, these assumptions are very constraining and unlikely to hold, and instead, an object can only observe the immediate reward after selecting and performing an action. The effective reinforcement learning is a key module of DECMDPs, which allows an agent to adapt to the dynamics of other agents and the context environment, and improves the system performance for cooperative multiagent systems. When an autonomous system is confronted with a sequence of MDPs chosen independently from a fixed distribution, their goal is to quickly find an optimal policy for each MDP. However, above those algorithm have no better way that shared any structure between MDPs.

Deisenroth et al. [12] developed a new multi-task policy search approach that provide an transferring knowledge framework for representing, learning, and reasoning about shared knowledge between MDPs. The key idea is to propose a parametrized policy as a function of both the state and the task, which allows learning a single policy that generalizes across multiple known and unknown tasks. This process can be expensive for even moderately large discrete environments, but is computationally intractable for the types of continuous, high-dimensional control problems considered here. Ammar et al. [13] proposed a multi-task policy gradient method to online learns a high-dimensional control policy based on previously learned knowledge with low computational overhead, transferring knowledge between tasks to accelerate learning. However, most contemporary multi-task policy gradient learning methods assume linear models and rely on Markov decision processes (MDPs).

Mnih [6] has proposed Deep Q Network, which can combine reinforcement learning with deep learning for sensory processing. Notably, recent advances in deep reinforcement

learning(DRL), in which several layers of nodes are used to build up progressively more abstract representations of the data, have made it possible for artificial neural networks to learn concepts such as object categories directly from raw sensory data. The deep reinforcement learning algorithm that is capable of human level performance on many Atari video games using unprocessed pixels for input. To do so, they use deep neural networks to approximate any of the following component of reinforcement learning: value function $V(s,\theta)$ or $Q(s,a,\theta)$, control policy $\pi(a|s;\theta)$, and sequential decision making model (state transition and reward). But the DRL algorithm can not learns sharing structure at every layer in a multi-task deep network. Yang [14] proposed two deep architectures which can be trained jointly on multiple related tasks. More specifically, we augment neural model with an external memory, which is shared by several tasks. Many tasks of interest, most notably physical control tasks, have continuous (real valued) and high dimensional action spaces, but those algorithm can only handle discrete and low-dimensional action spaces. Pfau [15] proposed an novel significant method that combines sequence generative adversarial networks for sensory processing with actor-critic methods. the method is alternative training methodology to generating sequential decision-making models, where the training procedure is a minimax game between a generative model and a discriminative model. This framework avoids the difficulty of maximum likelihood learning and has gained striking successes in natural image generation. However, little progress has been made in applying GANs to sequence temporal generation problems. This is due to the generator network in GAN is designed to be able to adjust the policy gradient continuously, which can not learn cross-task sharing structure at every layer in a deep network [16].

As pointed out connecting generative adversarial networks and deep reinforcement learning by Yang and Pfau, the cross-task high-dimensional sequential decision-making process in our paper can be formulated as a high-dimensional sequence generation, which can solve complex cross-domain coordination control from unprocessed, high-dimensional, sensory input. Our proposed MTGAN algorithm extends GANs with the multi-task policy learning, that a reward signal of task is provided by the critic network discriminator of our algorithm at the end of each episode via Monte Carlo approach, and the actor network generator of our algorithm captures the action network and learns the cooperative policy using estimated overall rewards.

## III. MULTI-TASK GENERATIVE ADVERSARIAL NETS FOR CROSS-DOMAIN DYNAMICAL COOPERATION

### A. Problem Formulation

To maximie coordinative efficiency in complex dynamical control network, the section assume a dynamical coordinating network consisting of $m$ finite number of autonomous machines. We propose the cross-domain coordination control framework(see Fig.1). The CDCC can consecutively learn control policy directly from raw sensory input of all of tasks, predict and assess system states, evaluate performance of system actions, make decisions, perform real-time optimization, and conduct fault diagnosis [17]. The aim of the CDCC is to address dynamical cooperation problem of large scale autonomous system on two different levels: critic network layer, and actor network layer. Each autonomous machine in the CDCC will run in a three-step closed loop: actor network generator $G_\theta$, critic network discriminator $D_\theta$, and selecting optimal joint action $Q^*(s,a)$. Our framework designs a multitask generative adversarial nets ( see Algorithms 1) to facilitate this loop in collaborative factory automation [18].

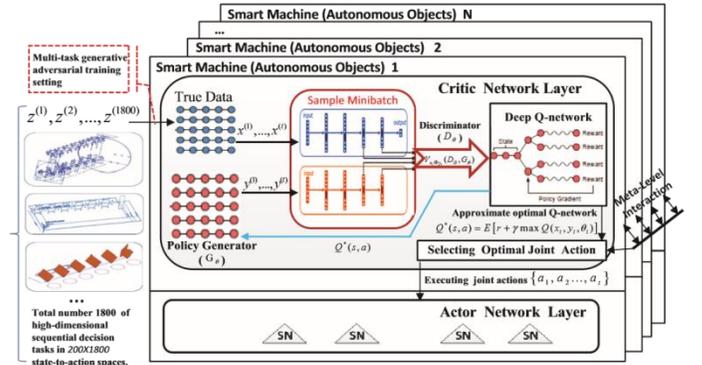

Fig. 1. The two-level distributed coordination control framework.

In the CDCC framework, autonomous machines are confronted with multiple high-dimensional consecutive tasks $Z^{(1)}, Z^{(2)}, \ldots, Z^{(T)}$ over its lifetime $t$. The Each $Z^{(t)} = (f(x_i^t, \theta_t), y_i^{(t)})$ is defined by a Sequential Decision Making (SDM) $f: x^{(t)} \to y^{(t)}$ from an observations state space $x^{(t)} \subseteq R^d$ to a action space $y^{(t)} \subseteq R^n$. The autonomous machine must learn successful directly policies $\theta^{(t)}$ from high-dimensional sensory inputs $x^{(t)}$, and generates multiple trajectories within each task before moving to the next new task. The tasks may be interleaved, providing autonomous machine the opportunity to revisit earlier tasks for experience, but the autonomous machine has no control over the task order.

We consider that machines are unaware of the total number of tasks $T$, the distribution of these high-dimensional tasks, or their order. When each machine receives a batch of $n_t$ high-dimensional sensory data from tasks $t$, the machine may be asked to make the correlations between the action-values $Q$ and the target values $r + \gamma \max Q(s,a)$. Its goal is to approximate the nonlinear action-value function $Q^*(s,a) = \max_\pi E[r_t + \gamma r_{t+1} + \gamma^2 r_{t+2} + \ldots | s_t = s, a_t = a, \pi]$, which is the maximum sum of reward $r_t$ discounted by $\gamma$ at each timestep $t$, archievable by a behaviour policy $\pi = P(a|s)$, after making an observation $(s)$ and taking an action $(a)$. When the total numbers of tasks $T_{max}$ and data instances $\sum_{t=1}^{T_{max}} n_t$ will be large, a long-standing challenge of our multi-task generative adversarial nets is to generate sequential deterministic policies directly from raw sensory input of all of tasks, and estimates the action-value function $Q^*(s,a)$ that a sample came from the training data rather than generative policy.

### B. Multi-Task Generative Adversarial Nets with Shared Memory

The section considers multi-task in which the autonomous machine interacts with the context of coordinating network through a sequence of observations, actions and rewards. The goal of the autonomous machine is to select joint actions in the multi-task environment that maximizes cumulative future reward. We define a action-value neural model $Q^*(s, a, \theta_t)$ with an policy parameters memory $\theta$, which is shared among all tasks. Our multi-task sequence generative adversarial nets maintains a library of $k$ latent action-value neural model $Q^*(s,a,\theta_t)$.

**Algorithm 1:** Multi-task generative adversarial nets with shared memory

1. **Initialize:** generate a sequence $Y_{1:T} = \{y_1, ..., y_T\}$.
2. **Initialize:** generator policy $G_\theta$, $D_\theta$ with random weights $\theta$.
3. **Initialize:** action-value function $Q(s_0, a_t)$ with random weights $\theta$.
4. **while** Received task $Z^{(t)}$ is available **do**
5.     $(X_{new}, Y_{new}, t) \leftarrow getNextTrainingData()$ ;
6.     $x_t \leftarrow X_{new}, y_t \leftarrow Y_{new}$ ;
7.     Generate a sequence $Y_{1:T} = \{y_1, ..., y_T\} \sim G_\theta$ ;
8.     **for** g-steps **do**
9.       Sample minibatch of $t$ examples $\{y_1, ..., y_t\}$ from data generating distribution $p_{data}(x)$ ;
10.       Update generator parameters via mult-task policy gradient $\theta \leftarrow \theta + \alpha_h \nabla_\theta \ell(\theta)$ ;
11.       Update generator $G_\theta$ via decending its stochastic gradient of Equation (6): $\nabla_{\theta_g} \frac{1}{T} \sum_{t=1}^{T} \lambda \log(D_\theta(Y_{1:T}))$;
12.     **end**
13.     **for** d-steps **do**
14.       Sample minibatch of $t$ noise samples $\{Z^{(1)}, ..., Z^{(t)}\}$ from noise prior $p_g(Z)$ ;
15.       Use current generator $G_\theta$ to generate negative examples $\{y_1, ..., y_T\}$ and combine with given true data $(x_t, y_t, \theta)$ ;
16.       Alternatively training discriminator $D_\theta$ via aecending its stochastic gradient of Equation (7): $\nabla_{\theta_d} \frac{1}{T} \sum_{t=1}^{T} [\mu \log(1 - D_\theta(G_\theta(y_t))) + \lambda \log(D_\theta(Y_{1:T}))]$;
17.     **end**
18.     $t \leftarrow t+1$ ;
19.     Compute $Q(x_t, y_t, \theta_t)$ by Equation (5) ;
20. **end**
21. **Return** $Q^{\theta_t}(x_t, y_t)$ ;

When each machine perform experience from each tasks environment, the $G_\theta$ produce a sequential decision sequence $Y_{1:T} = (y_1,...,y_t,...,y_T)$ using Long Short-Term Memory(LTSM) [19] at time-step $t$. The $s$ of $Q^*(s, a, \theta_t)$ is the current produced sequence $(y_1,...,y_{t-1})$ and the action $a$ is the next sequence $y_t$ to select. At the same time, the Q-network discriminator $D_\theta$ is trained by providing positive examples from the real sequence data and negative examples from policy generator $G_\theta$. The generator $G_\theta$ is updated simultaneously by deterministic policy gradient (DPG) and end reward received from the $D_\theta$.

The $\theta_t$-parameterized of single-task update at iteration $t$ using following deterministic policy gradient function:

$$\begin{aligned} J(\theta_t) &= \int_S \rho_{\theta_t}(s_0) \int_A \pi_{\theta_t}(x_t, y_t) Q(x_t, y_t, \theta_t) da ds \\ &= \mathbb{E}_{s \sim p_{\theta_t}, a \sim \pi_{\theta_t}}[r(x_t, y_t)] \\ &= \sum_{y_t \in Y} G_\theta(y_t|Y_{1:T}) \cdot Q_{D_\theta}^{G_\theta}(x_t, y_t), \end{aligned}$$

which $Q_{D_\theta}^{G_\theta}(x_t, y_t)$ is the action-value function of a sequence $Y_{1:T}$, which in starting from state $s$, taking action $a$, and then following policy $G_\theta$. We estimate the optimal action value function $Q_{D_\theta}^{G_\theta}(a = yT, s = Y_{1:T-1})$ via an adversarial process, in which we simultaneously train two models: a generator $G_\theta(y_t|Y_{1:T})$ that generates a decision sequence, and a discriminator $D_\theta(Y_{1:T}^n)$ estimates a sequence probability that is from the true data rather than $G_\theta(y_t|Y_{1:T})$. The training procedure for $G_\theta(y_t|Y_{1:T})$ is to maximize the probability of by minimax two-player game with value function $Q_{D_\theta}^{G_\theta}(a = Y_t, s = Y_{1:T-1})$.

The $Q_{D_\theta}^{G_\theta}(a = Y_t, s = Y_{1:T-1})$ has capability to store long term information and knowledge shared by several related tasks. This mult-tasks learning problem is realized by the objective function based on efficient lifelong learning algorithm (ELLA) [20]:

$$e_T(Q^*) = \frac{1}{T} \sum_{t=1}^{T} \min_{G_\theta}[J(\theta_t) + \mu \parallel G_\theta(y_t|Y_{1:T}) \parallel_1] \\ + \lambda \|Q^{\theta_t}(x_t, y_t)\|_F^2, \quad (1)$$

which the $(x_t, y_t)$ is the $i$-th training instrance for task $t$, and $J(\theta_t)$ is a loss function for updating Q-network at iteration $i$. The $L_1$ of $G_\theta(y_t|Y_{1:T})$ is used as a convex approximation to the true vector sparsity. The $\|.\|_z$ is the frobenius norm with the $\theta$-parameterized for transferring share neural model across tasks. The $\mu$ denotes task weights of generator for each task. The $\lambda$ denotes weights of discriminator model for each task. However, the Equation (1) is not jointly train in $G_\theta(y_t|Y_{1:T})$ and $D_\theta(Y_{1:T}^n)$, due to two inefficiencies: a) the eliminating dependence of $G_\theta(y_t|Y_{1:T})$ for all tasks, and b) how to alternately train the discriminator $D_\theta(Y_{1:T}^n)$ from large-scale state spaces $R^d$.

### C. Eliminating Dependence of Actor Network Generator

To eliminate the first inefficiency, we approximate Equation (1) in data spaces vector $Z^t$ by the second-order Taylor expansion of $\ell(\theta^t)$ around $\theta^t$. Plugging the second-order Taylor expansion into Equation (1) yields:

$$e_T(Q^*) = \frac{1}{T} \sum_{t=1}^{T} \min_{G_\theta}[\| \theta^t - Q^{\theta_t}(x_t, y_t) G_\theta(y_t|Y_{1:T}) \|_{Z^t}^2 \\ + \mu \parallel G_\theta(y_t|Y_{1:T}) \parallel_1] + \lambda \|Q^{\theta_t}(x_t, y_t)\|_F^2, \quad (2)$$

Which:

$$\begin{aligned} \theta^t &= \nabla_{\theta^t} J(\theta) \\ &= \mathbb{E}_{y_t|Y_{1:t-1} \sim G_\theta}[\nabla_{\theta^t} G_\theta(y_t|Y_{1:t-1}^n) \cdot D_\theta(Y_{1:t}^n)] \\ &= \mathbb{E}_{y_t|Y_{1:t-1} \sim G_\theta}[\nabla_{\theta^t} \log D_\theta(Y_{1:t}^n)], \quad (3) \end{aligned}$$

$$\begin{aligned}Z^t &= \nabla^2_{(\theta^t,\theta^t)} J(\theta) \\ &= \mathbb{E}_{y_t|Y_{1:t-1}\sim G_\theta}[\nabla^2_{(\theta^t,\theta^t)} G_\theta(y_t|Y^n_{1:t-1}).D_\theta(Y^n_{1:t})] \\ &= \mathbb{E}_{y_t|Y_{1:t-1}\sim G_\theta}[\nabla^2_{(\theta^t,\theta^t)} \log(1-D_\theta(G_\theta(z^t)))], \quad (4)\end{aligned}$$

$$Q^{\theta_t}(x_t = Y_{1:T}, y_t = y_T) = D_\theta(Y_{1:T}), \quad (5)$$

We train a sequential policy generator $G_\theta(y_t|Y_{1:t-1})$ directly from raw sensory data through Long Short Term Memory(LSTM) with global shared memory, and produce a sequence $Y_{1:T} = (y_1,...,y_t,...,y_T)$, $y_t \in Y$. The $G_\theta(y_t|Y_{1:t-1})$ maps the input embedding representations $x_1,...,x_t$ into a sequence of hidden states $\sim_1,...,\sim_t$ by the update function $g(\sim_{t-1},x_t)$ recursively. Moreover, a softmax output layer z maps the hidden states into output token distribution $G_\theta(y_t|Y_{1:t-1}) = softmax(c+V\sim_t)$, in which the parameters are a shared latent basis vector $c$ and a weight matrix $V$. The softmax mechanism can be used as a generator in multi-task sequence generative adversarial nets.

We incorporate the maximum likelihood estimation (MLE) in our objective function, Rewriting the error term in Equation (1) in terms of the lower bound yields:

$$e_T(Q^*) = \frac{1}{T}\sum_{t=1}^{T}\mathbb{E}_{y_t|Y_{1:t-1}\sim G_\theta}[\lambda \log(D_\theta(Y_{1:T}))], \quad (6)$$

where the E[.] can be approximated by stochastic policy sampling methods [21], we update the parameters $\theta$ of generator by $\theta + \alpha_h \nabla_{\theta_t} \ell(\theta)$, in which $\alpha_h \in R^+$ denotes the corresponding learning rate at $h$-th step. Simultaneously, the generative model $G_\theta(y_t|Y_{1:t-1})$ is updated by employing a policy gradient and Monte Carlo search [22] with a roll-out policy $G_\theta$ from the discriminative model $D_\theta$. The reward is estimated by the likelihood that it would fool the discriminative model $D_\theta$.

### D. Alternately Training Critic Network Discriminator

In order to solve the second inefficiency of Equation (1), the section is to alternately train $D_\theta(G_\theta(y_t))$ following two-player minimax game with $Q^{G_\theta}_{D_\theta}(a = y_T, s = Y_{1:T-1})$:

$$e_T(Q^*) = \frac{1}{T}\sum_{t=1}^{T}\mathbb{E}_{Y_{1:t-1}\sim G_\theta}[\mu \log(1 - D_\theta(G_\theta(y_t|Y_{1:T}))) + \lambda \log(D_\theta(Y_{1:T}))], \quad (7)$$

which $D_\theta(Y^n_{1:t})$ is train by the Convolutional Neural Network(CNN) while holding the $G_\theta$ re-trained, and another in the $G_\theta$ is optimized while holding $D_\theta$ co-training. These two steps are then repeated until convergence. In order to overcome collapse problem, the number of negative examples from $G_\theta(Y_{1:t})$ is the same as the positive examples in each d-step and g-step, where the generator and discriminator are trained simultaneously in Algorithms 1.

When training the discriminator $D_\theta(Y_{1:t})$ by positive examples and negative examples, we first prepresent an input sequence $x_1,...,x_T$ of CNN as:

$$Y_{1:t} = x_1 \oplus x_2 \oplus ... \oplus x_t, \quad (8)$$

which $x_t \in R^k$ is the k-dimensional token embedding and $\oplus$ is the concatenation operator to build the matrix $Y_{1:t} \in R^{T \times k}$. Then a kernel $\omega \in R^{l \times k}$ applies a convolutional operation to a window size of l to produce a new feature map:

$$D_\theta(Y_{1:t}) = \rho(\omega \oplus Y_{1:t} + b), \quad (9)$$

where $\oplus$ operator is the summation of element production, b is a bias term and $\rho$ is a non-linear function. We then use various numbers of kernels with different window sizes to extract different features.

Finally, we apply a max-over-time pooling operation on the feature map $\tilde{D}_\theta(Y_{1:t}) = \max\{\rho_1,...,\rho_{T-l+1}\}$ and pass all pooled features from different kernels to a fully connected softmax layer to get the probability that a given sequence is real. We use different sets of negative samples combined with positive ones for reducing the variability of the estimation.

## IV. EXPERIMENTS AND ANALYSIS

### A. Experimental Scenarios Setting

In this section, we draw up three experimental scenarios to evaluate the effectiveness and efficiency of our proposed multitask dynamical control framework and algorithms based on factory dataset of Weichai Power company(See Fig.2), which is well-known for providing engine equipment for heavy trucks. Our multi-task dynamical control experimental platform is implemented by AFME [23] and Google TensorFlow with algorithm 1 in Section III.

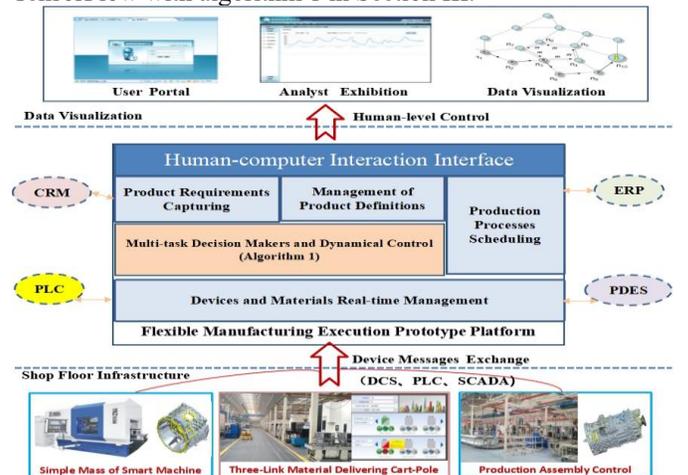

Fig. 2. The prototype system of complex dynamical control

The prototype system has been deployed within China Weichai Power Flexible Manufacturing Platform, where has twenty Intel Xeon E5-4600 server with four 8-core and 64 GB memory and runs the linux (kernel 2.6.8) operating system. In addition to enable the connection of 300 smart devices to the gateway it supports real-time ethernet network connections. The experimental platform has capability of processing 100TB data sets from a flexible manufacturing line consisting of 50 smart machines, which record their running status and manufacturing process data every 20 millisecond. This flexible manufacturing line supports 500 varieties of engine parts manufacturing tasks and has the capability to assemble simultaneously eight types of engine. This paper focuses on transfer both between tasks in the same domain as well as between tasks in different domains.

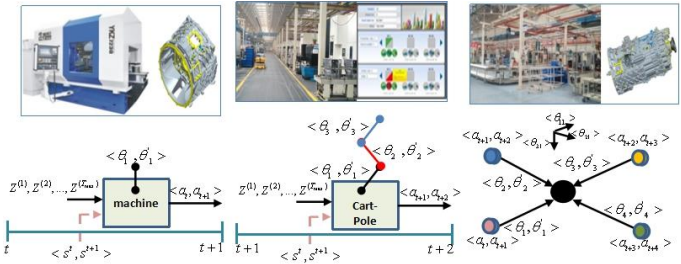

Fig. 3.  Three flexible manufacturing control systems in the experiments

We applied multi-task generative adversarial nets with shared memory algorithm to learn control policies for the three dynamical system shown in Fig.3. The Multi-task Dynamical Control module received 50 varieties of manufacturing tasks from Management of Product Definitions System(MPDS), yielding a set of tasks from each domain with varying dynamics. We evaluated multi-task generative adversarial nets with shared memory algorithm on three manufacturing dynamic control systems. In each control system, the distance between the current state and the goal position was used as the reward function [24]. On all systems, the reward function is based on two factors: a) penalizing states far from the goal state, and b) penalizing actions to encourage smooth, low-energy manufacturing.

Simple Mass of Smart Machine(SMSM): The goal with the SMSM is to control machining technology of engine shell structure according on specified shell blue print. The dynamical system is characterized by four parameters: two running states variables $\langle s^t, s^{t+1} \rangle$, one control policy $\langle \theta_1, \theta_1' \rangle$, action space $\langle a^t, a^{t+1} \rangle$, and transition probability function $P$ for dynamics of system. In our experiments, the goal of SMSM is to design a measuring processing performance of machine $\pi_\theta(a_t|s_t) \to 7 [0,1]$ from $\langle s_0^t, s_1^t \rangle$ to $\langle s_0^{t+1}, s_1^{t+1} \rangle$, where number of tasks $t \in \{1,2,...,50\}$.

Three-Link Material Delivering Cart-Pole(TLMDCP): The TLMDCP is a highly nonlinear and difficult system to control. The TLMDCP system dynamics are described via an eight-dimensional state vector $\langle s, \dot{s}, \vartheta_1, \dot{\vartheta}_1, \vartheta_2, \dot{\vartheta}_2, \vartheta_3, \dot{\vartheta}_3 \rangle$, where $s$ and $\dot{s}$ describe the position and velocity of the cart and $\vartheta_t$ and $\dot{\vartheta}_t$ represent the angle and angular velocity of the $t$th link. The system is controlled by applying a force $\langle a^t, a^{t+1} \rangle$ to the cart in the $\langle s^t, s^{t+1} \rangle$ direction, with the goal of balancing the three poles upright.

Production Assembly Control(PAC): The system dynamics were adopted from a simulator validated on real quadrotors [23], and are described via three angles and three angular velocities in the body frame $\langle \vartheta_1, \vartheta_2, \vartheta_3, \vartheta_4, \vartheta_{11}, \vartheta_{21}, \vartheta_{31} \rangle$. The actions consist of four rotor torques $\langle a_t, a_{t+1}, a_{t+2}, a_{t+3}, a_{t+4} \rangle$. Each task corresponds to a different configuration (e.g., different armature lengths,etc.) and the goal is to stabilize the different controller [24].

B. *Our Algorithm Generating Sequential Decision Policy in Simple Mass of Smart Machine Domain*

The experiment first gave 50 engine manufacturing parts high-dimensional 3D data (84 × 84 × 4) from MPDS, where t ∈ [1,...,50], batch-size = 25, learning-rate = 0.1, gamma = 0.99 and $\gamma \in [0,1]$. Each instance of high-dimensional 3D data is a monochrome image of size 84 × 84 × 4, and These parameters were chosen to ensure a variety of tasks, including the set of states and possible actions $\langle s_t, a_t, \theta_t \rangle$ which were dynamically controlled with SMSM. Then, we use LSTM network to generate 10000 sequences of length 20 as the training set S for the generative models, updating $G\theta(y_t|Y_{1:t-1})$ and the $Q_{D_\theta}^{G_\theta}(x_t, y_t)$ after each session. During the training process, the multi-task generative adversarial nets with shared memory algorithm was limited to eighty trajectories for SMSM within 150 time steps each to perform the update. our goal is to generate a set of optimal control policies $\{\pi\theta(1),...,\pi\theta(T)\}$ from 50 engine manufacturing parts high-dimensional 3D data.

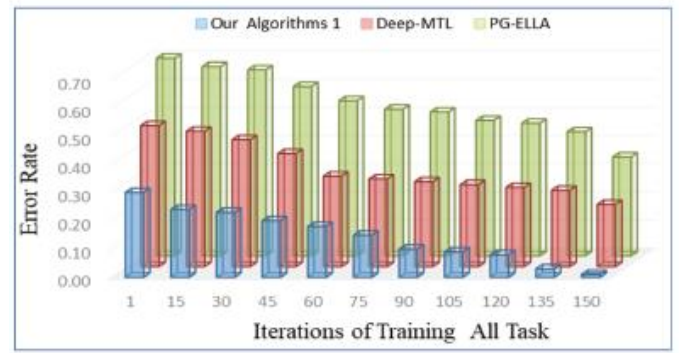

Fig. 4. Our proposed algorithm generate control policies for Simple Mass of Smart Machine Scenario based on devising various engine parts 3D data. Discuss error rate $\gamma$ of our Algorithms 1,Deep-MTL, and PG-ELLA within 150 time steps each to perform the update.

Fig.4 compares our Algorithms 1 to Deep-MTL, and PGELLA, showing the error rate of generative share models for each task versus the number of training iterations. Our Algorithms 1 clearly outperforms Deep-MTL, and PG-ELLA in both the initial and final performance on all high-dimensional task domains, and demonstrating best user-defined MTL when the training data is very small, and their performance is comparable when the training data is large. However DeepMTL does not consistently improve on STL, and even reduces performance when the training set is bigger, and the PG-ELLA is limitations when the goal is for generating sequences of discrete tokens. This difference in our Algorithms 1 is attributed to sufficient data eventually providing some effective taskspecific representation from high-dimensional training data.

### C. The Accuracy Rate of Our Algorithm Generating Sequential Decision Policy in Production Assembly Control Domain

In this experiment, the production assembly control with dynamics is influenced by inertial constants around $911, 921, 931$, how control production assembly using overall variation of the system state, and the length of the rods supporting nine rotors.

We focus on stability of production assembly and so consider only nine of those state variables. The production assembly system has a high-dimensional action space, where the goal is control the four precision casting $\{\theta_i\}_{i=1}^4$ of mechanical transmission to stabilize the system. We first fit together 15 engine manufacturing parts by varying each of: the X-axix velocities $\theta_{xx} \in [120+\lambda, 160+\lambda]$, the Y-axix velocities $\theta_{yy} \in [220+\lambda, 260+\lambda]$, the Z-axix velocities $\theta_{zz} \in [40+\lambda, 70+\lambda]$. We transform the 15 engine manufacturing parts into sequences numbers of production assembly line from 1 to 200 with the depth 32. During the training process, the multi-task generative adversarial nets with shared memory algorithm was limited to 50 trajectories for PAC within 150 time steps each to perform the update. our goal is to generate a set of optimal production assembly control policies.

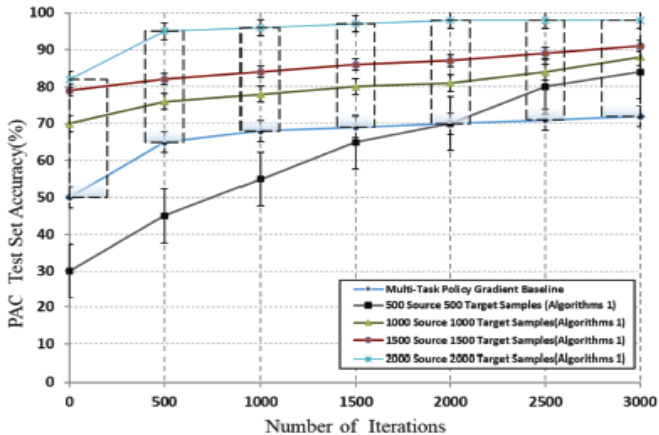

Fig. 5. The accuracy rate of generating sequential decision policy by multitask generative adversarial nets with shared memory algorithm in PAC domain. Discuss the optimal action-value reward without systematic biases between our Algorithms 1 and multi-task policy gradients baseline within 150 time steps each to perform the update.

Fig.5 compares our Algorithms 1 to PG-ELLA on production assembly control scenario. We set point multi-task policy gradients baseline, which presents mean return from a naive policy which samples actions from a uniform distribution over the valid action space. As on the production assembly systems, we see that our Algorithms 1 clearly outperforms PG-ELLA in both the initial and final performance, and this more best performance increases as our Algorithms 1 learns more 2000 targets of task. The accuracy rate of the policy learned by our Algorithms 1 after observing all tasks is significantly better than the policy learned using PG-ELLA, showing the benefits of knowledge transfer between tasks. Most importantly for discrete-time nonlinear applications, by using the $G_\theta(y_t|Y_{1:t-1})$ and $D_\theta(Y_{1:t})$ alternately training over previous tasks, our Algorithms 1 can achieve high performance in a new task much more quickly (with fewer trajectories) than PG-ELLA within 150 time steps. Our Algorithms 1 extends end-to-end reinforcement learning that uses reward to continuously shape representations within the convolutional network towards target features of the environment that facilitate actionvalue function convergence very quickly. we used an LSTM iterative update that adjusts the action-values $Q$ towards target values that are only periodically updated, thereby reducing correlations with the target.

### D. The Accuracy Rate of Our Algorithm Generating Sequential Decision Policy from SMSM Domain to PAC Domain

In this experiment, we considered the more difficult problem of cross-task sharing policy structure transfer. The experimental setup is identical to the same-domain case with the crucial difference that the state and action spaces were different for the source and the target task (since the tasks were from different domains). We tested a transfer scenarios of cross-domain tasks from SMSM to PAC. In each experimental case, the source and target task have different numbers of state variables and system dynamics. The source (from 500 to 2000) and target (from 500 to 2000) tasks are improved using the multi-task generative adversarial nets with shared memory algorithm.

We set point multi-task policy gradients baseline, which presents mean return from a naive policy which samples actions from a uniform distribution over the valid action space. Fig.6 shows the results of cross-domain transfer, demonstrating that the multi-task generative adversarial nets with shared memory algorithm can achieve the successful transfer of tasks from SMSM domain to PAC domain. These results show that a) transfer-initialized policies outperform multi-task policy gradients, even between different flexible manufacturing domains and b) initial and final performance improves as more samples are used to learn optimal policies $\Pi*$. The experimental result shows that the multi-task generative adversarial nets with shared memory algorithm is capable of: a) automatically generating inter-task mapping $\pi_\vartheta(a|s)$ to facilitate knowledge transfer across domains and b) effectively transferring from SMSM to PAC. When the source and target tasks are highly dissimilar, the multi-task generative adversarial nets with shared memory algorithm is capable of successfully providing target policy initializations that outperform state-of-the-art multi-task policy gradient techniques.

making in three discrete-time nonlinear systems (SMSM, TLMDCP, PAC) with flexible manufacturing.

Three experimental results show important value for improving current collaborative factory automation. Our future work will focus on B-GAN in the model-free adaptive control, model-free adaptive iterative learning control, and improving the descriptions to estimate dynamic behavior for collaborative factory automation.

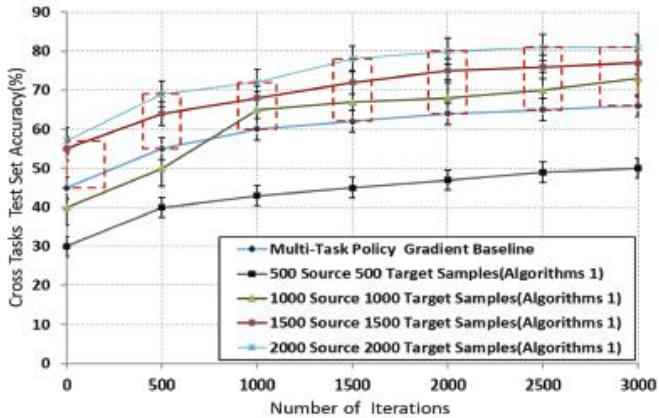

Fig. 6. The accuracy rate of generating sequential decision policy by multitask generative adversarial nets with shared memory algorithm from SMSM domain to PAC domain. Discuss the optimal action-value reward without systematic biases between our Algorithms 1 and multi-task policy gradients baseline within 150 time steps each to perform the update.

## V. CONCLUSION

In this paper, we have described a novel cross-domain coordination control framework for solving collaborative factory automation from unprocessed, high-dimensional, sensory input by multi-task generative adversarial nets. Our experimental results show the framework can efficiently learn control policy directly from raw sensory input of all of tasks, predict and assess system states, evaluate performance of system actions, and perform real-time optimization in collaborative factory automation, while huge amounts of online or offline measured process data can easily be collected by the well-developed information technology.

More importantly, multi-task generative adversarial nets with shared memory has been developed which can simultaneously training actor network generator and critic network discriminator from collaborative factory automation across task domain scenario. The actor network generator produces an action coordination network given the current state of the task environment, and the critic network discriminator that estimates the action-value function from the training data rather than actor network generator. The alternatively training make each autonomous system run in an iterative three-step closed loop of framework. Additionally, we discussed those performance of our algorithms online generating and estimating sequential decision

## ACKNOWLEDGMENT

The authors also would like to thank anonymous editor and reviewers who gave valuable suggestion that has helped to improve the quality of the manuscript. This research has been supported by the Project for Natural Science Foundation of China No.$U1636220$, No.$61772525$.